\documentclass[final,12pt]{msml2022} 

\title[An Upper Limit in Linear Frequency Principle Model]{An Upper Limit of Decaying Rate with Respect to Frequency in Linear Frequency Principle Model}

\usepackage{times}                                      %
\usepackage{mlmath}                                     %
\usepackage{graphicx}                                   %
\usepackage{subfloat}                                   %
\usepackage{caption}                                    %
\usepackage{amsmath,amsfonts,amssymb,mathtools}         %
\newtheorem{thm}{Theorem}                               %
\newtheorem{lem}{Lemma}                                 %
\newtheorem{prop}{Proposition}                          %
\newtheorem{rmk}{Remark}                                %
\newtheorem{exam}{Example}                              %
\newtheorem{prob}{Problem}                              %
\usepackage{color}                                      %
\msmlauthor{
    \Name{Tao Luo}\textsuperscript{\rm 1,2}\thanks{Corresponding author} \Email{luotao41@sjtu.edu.cn}
    \AND
    \Name{Zheng Ma}\textsuperscript{\rm 1,2} \Email{zhengma@sjtu.edu.cn}
    \AND
    \Name{Zhiwei Wang}\textsuperscript{\rm 1}\thanks{PhD student of School of Mathematical Sciences at Shanghai Jiao Tong University.} \Email{victorywzw@sjtu.edu.cn}
    \AND
    \Name{Zhi-Qin John Xu}\textsuperscript{\rm 1} \Email{xuzhiqin@sjtu.edu.cn}
    \AND
    \Name{Yaoyu Zhang}\textsuperscript{\rm 1,3} \Email{zhyy.sjtu@sjtu.edu.cn}\\
    \addr \textsuperscript{\rm 1}School of Mathematical Sciences, Institute of Natural Sciences, MOE-LSC and Qing Yuan Research Institute,\\
    Shanghai Jiao Tong University, Shanghai, 200240, P.R. China \\
    \textsuperscript{\rm 2} CMA-Shanghai\\
    \textsuperscript{\rm 3}
    Shanghai Center for Brain Science and Brain-Inspired Technology
}

\makeatletter
    \let\Ginclude@graphics\@org@Ginclude@graphics
\makeatother

\begin{document}

\maketitle

\begin{abstract}%
    Deep neural network (DNN)  usually learns the target function from low to high frequency, which is called frequency principle or spectral bias. This frequency principle sheds light on a high-frequency curse of DNNs --- difficult to learn high-frequency information. Inspired by the frequency principle, a series of works are devoted to develop algorithms for overcoming the high-frequency curse. A natural question arises: what is the upper limit of the decaying rate  w.r.t. frequency when one trains a DNN? 
    In this work, we abstract a paradigm for modeling and analysis of algorithm suitable for supervised learning problems from the sufficiently wide two-layer neural network, i.e., linear frequency principle (LFP) model. Our theory confirms that there is a critical decaying rate w.r.t. frequency in LFP model. 
    It is precisely because of the existence of this limit that a sufficiently wide DNN interpolates the training data by a function with a certain regularity. However, if the decay rate of an algrithm in our paradigm is above such upper limit, then this algorithm interpolates the training data by a trivial function, i.e., a function is only non-zero at training data points. This work rigorously proves that the high-frequency curse is an intrinsic difficulty of the LFP model, which provides similar insight to DNN.
\end{abstract}


\section{Introduction\label{sec:Introduction}}
        
    The study of generalization in deep learning attracts much attention in recent years due to the contradiction to the traditional wisdom \citep{breiman1995reflections,zhang2016understanding}, that is, over-parameterized DNNs often generalize well in real dataset. Recently, a series of works have demonstrated a frequency principle (F-Principle), an implicit bias in Fourier domain, that is, a DNN tends to learn a target function from low to high frequencies during the training~\citep{xu_training_2018,xu2019frequency,rahaman2018spectral}. \cite{xu2018understanding,xu2019frequency} propose that the low-frequency bias is due to that a function with a certain regularity decays w.r.t. frequency in the Fourier domain with a certain rate. This mechanism is further confirmed by a series of theoretical works  \citep{luo2019theory,zhang_explicitizing_2019,Luotao2020LFP,cao2019towards,yang2019fine,basri2019convergence,bordelon2020spectrum,e2019machine}. The frequency principle implies a rational that DNNs  generalize well for real datasets, which are often low-frequency dominant \citep{xu2019frequency}. Meanwhile, such low-frequency bias also suggests a high-frequency curse, i.e., DNNs are difficult to learn high-frequency information. To overcome the high-frequency curse, various approaches are proposed \citep{xu2019frequency,jagtap2019adaptive,biland2019frequency,cai2019phasednn,peng2020accelerating,cai2019multi,liu2020multi,li2020multi,wang2020multi,tancik2020fourier,mildenhall2020nerf,agarwal2020neural,campo2020band,jiang2020focal,xi2020drl}.
    
    A natural question is that what is the upper limit of the decaying rate w.r.t. frequency in DNN. The decaying rate means the decaying speed of the Fourier coefficient of the difference of the neural network function and the target function with respect to the frequency. Such an upper limit characterizes the boundary of the frequency bias, providing a better understanding of the implicit bias of DNNs in Fourier domain. In addition, it also provides a guidance for algorithm design of DNNs which could be more efficient in capturing high-frequency information. 
    In this work, we use Linear Frequency Principle (LFP) \citep{Luotao2020LFP} dynamics to model the dynamical behavior of a sufficiently wide two-layer neural network in the linear regime \citet{jacot2018neural}, and prove that there is a critical decaying rate w.r.t. frequency in LFP model. 
    It is precisely because of the existence of this limit that the LFP model interpolates the training data by a function with a certain regularity. However, if the decay rate of an algrithm in our paradigm is above such upper limit, then this algorithm interpolates the training data by a trivial function, i.e., a function is only non-zero at training data points.
    
    Theoretical works have estimated the decaying rate w.r.t. frequency follows a power law for DNNs with a certain regularity activation function in the gradient descent training \citep{luo2019theory,zhang_explicitizing_2019,Luotao2020LFP,cao2019towards,basri2019convergence,bordelon2020spectrum}. The long-time limit solution of such gradient descent training is proved to be equivalent to solving a Fourier-domain variational problem \citep{zhang_explicitizing_2019,Luotao2020LFP}. Inspired by above works about the F-Principle, in this paper, we propose a general Fourier-domain variational formulation for supervised learning problem, including DNNs in linear regime, and study its well-posedness. In continuum modelling, it is often difficult to impose the constraint of given values on isolated data points in a function space without sufficient regularity, e.g., a $L^p$ space. We circumvent this difficulty by regarding the Fourier-domain variation as the primal problem and the constraint of isolated data points is imposed through a linear operator. Under a necessary and sufficient condition within our unified framework, we establish the well-posedness of the Fourier-domain variational problem. We show that the well-posedness depends on a critical exponent, which equals to the data dimension. This is a stark difference compared with a traditional partial differential equation (PDE) problem. For example, in a boundary value problem of any PDE in a $d$-dimensional domain, the boundary data should be prescribed on the $(d-1)$-dimensional boundary of the domain, where the dimension $d$ plays an important role. However, in a well-posed supervised learning problem, the constraint is always on isolated points, which are $0$-dimensional independent of $d$, while the model has to satisfy a well-posedness condition depending on the dimension. In practice, common DNNs in linear regime is a convenient way to implement our formulation. Therefore, the convergence rate of high-frequency has a upper limit. An algorithm with too fast high-frequency learning would lead to a learned function only non-zero at training data points. Such understanding of the upper limit of decaying rate  indicates a better way to overcome the high-frequency curse is to design a proper pre-condition approach to shift high-frequency information to low-frequency one, which coincides with several previous developed algorithms for fast learning high-frequency information \citep{cai2019phasednn,cai2019multi,liu2020multi,li2020multi,wang2020multi,tancik2020fourier,mildenhall2020nerf,agarwal2020neural}.
    
    The rest of the paper is organized as follows. Section~\ref{section2} shows some related work. In section~\ref{section3}, we propose a Fourier-domain variational formulation for supervised learning problems. The necessary and sufficient condition for the well-posedness of our model is presented in section~\ref{section4}. Section~\ref{section5} is devoted to the numerical demonstration in which we solve the Fourier-domain variational problem using band-limited functions. Finally, we present a short conclusion and discussion in section~\ref{sec:conclusion}.

\section{Related Works}\label{section2}
    It has been an important approach to study machine learning from the perspective of implicit bias \citep{neyshabur2014search}, such as the implicit bias of training algorithms \citep{gunasekar2018characterizing,soudry2018implicit}, dropout \citep{mianjy2018implicit}, linear network \citep{gunasekar2018implicit} and DNNs under different initializations  \citep{luo2020phase}.

    The low-frequency implicit bias is named as frequency principle (F-Principle) \citep{xu_training_2018,xu2019frequency} or spectral bias \citep{rahaman2018spectral} and can be robustly observed no matter how overparameterized NNs are.  \cite{xu2018understanding,xu2019frequency} propose a key mechanism of the F-Principle that the regularity of the activation function  converts into the decay rate of a loss function in the frequency domain.  Theoretical studies subsequently show that the F-Principle holds in general setting with infinite samples \citep{luo2019theory} and in the regime of wide NNs (Neural Tangent Kernel (NTK) regime \citep{jacot2018neural}) with finite samples \citep{zhang_explicitizing_2019,Luotao2020LFP} or samples distributed uniformly on sphere \citep{cao2019towards,yang2019fine,basri2019convergence,bordelon2020spectrum}. \citet{e2019machine} show that the integral equation would naturally leads to the F-Principle. In addition to characterizing the training speed of DNNs, the F-Principle also implicates that DNNs prefer low-frequency function and generalize well for low-frequency functions \citep{xu2019frequency,zhang_explicitizing_2019,Luotao2020LFP}.

\section{Notations}
    In the following, we consider the regression problem of fitting a target function $f^{*}\in C_c(\sR^d)$. Clearly, $f^{*}\in L^{2}(\sR^d)$. Specifically,
    we use a sufficiently wide two-layer DNN, $h(\vx,\vtheta(t))$ with a parameter set $\vtheta(t)$ which varies with the training time $t$,
    to fit the training dataset $\{(\vx_{i},y_{i})\}_{i=1}^{n}$ of $n$ sample
    points, where $\vx_{i}\in\sR^d$, $y_{i}=f^{*}(\vx_{i})$ for each $i$.
    For the convenience of notation, we denote $\vX=(\vx_{1},\ldots,\vx_{n})^\T$,
    $\vY=(y_{1},\ldots,y_{n})^\T$.

    It has been shown in \cite{jacot2018neural,lee2019wide} that, if the number of neurons in each hidden layer is sufficiently large, then $\norm{\vtheta(t)-\vtheta(0)}\ll1$ for any $t\ge 0$. In such cases, $h(\vx,\vtheta(t))$ can be approximated up to the first order in the Taylor expansion at $\theta(0)$, which means the the following function
    \begin{equation}
        h_{\mathrm{lin}}(\vx,\vtheta)=h(\vx,\vtheta_{0})+\nabla_{\vtheta}h\left(\vx,\vtheta_0\right)\cdot(\vtheta-\vtheta_{0}),\label{eq:linear}
    \end{equation}
    is a very good approximation of DNN output $h(\vx,\vtheta(t))$
    with $\vtheta(0)=\vtheta_{0}$. Note that,
    we have the following requirement for $h$ which is
    easily satisfied for common DNNs: for the gradient in equation (\ref{eq:linear}), we require that for any $\vtheta\in\sR^{m}$,
    there exists a weak derivative of $h(\cdot,\vtheta_0)$
    with respect to $\vtheta$ satisfying $\nabla_{\vtheta}h(\cdot,\vtheta_0)\in L^{2}(\sR^d)$.
    
    A two-layer neural network is 
    \begin{equation}
        h(\vx,\vtheta(t))=\sum_{j=1}^{m} a_j \sigma (\vw_j\cdot \vx + b_j),
    \end{equation}
    where $\sigma$ is the activation function, $\vw_j$ is the input weight, $a_j$ is the output weight, $b_j$ is the bias term. 
    
    In this work, for any function $g$ defined on $\sR^d$, we use the following convention of the Fourier transform and its inverse:
    \begin{equation*}
        \fF[g](\vxi)=\int_{\sR^d}g(\vx)\E^{-2\pi \I\vxi^\T\vx}\diff\vx,\quad g(\vx)=\int_{\sR^d}\fF[g](\vxi)\E^{2\pi \I\vx^\T\vxi}\diff\vxi.
    \end{equation*}

\section{Fourier-domain Variational Problem for Supervised Learning}\label{section3}
    To study the decaying rate limit w.r.t. frequency in DNN training, we propose a Fourier-domain variational problem for supervised learning, in which frequency bias can be imposed by weight term $\langle\vxi\rangle^\alpha$ in equation (\ref{mini-prob}). To show the motivation and the rationality of the variational problem, we first introduce a linear frequency principle.  
    \subsection{Motivation: Linear Frequency Principle}
        In the large width limit, it is reasonable \citep{jacot2018neural,lee2019wide} to assume a linear condition, i.e.,  $h(\vx,\vtheta)=h_{\mathrm{lin}}(\vx,\vtheta)$. Based on the linear condition, 
        \cite{zhang_explicitizing_2019,Luotao2020LFP} derived a Linear F-Principle (LFP)
        dynamics to effectively study the training dynamics of a two-layer NN with the mean square loss in the large width limit. Up to a multiplicative constant
        in the time scale, the gradient
        descent dynamics of a sufficiently wide two-layer NN is approximated by
        \begin{equation}
            \partial_{t}\fF[u](\vxi,t)=-\,(\gamma(\vxi))^{2}\fF[u_{\rho}](\vxi),\label{eq:ReLUnnFP}
        \end{equation}
        \noindent where $u(\vx,t)=h(\vx,\vtheta(t))-f^{*}(\vx)$, $u_{\rho}(\vx,t)=u(\vx,t)\rho(\vx)$,  $\rho(\vx)=\frac{1}{n}\sum_{i=1}^n\delta(\vx-\vx_i)$, accounting for the real case of a finite training dataset $\{(\vx_i,y_i)\}_{i=1}^n$, and $\gamma(\vxi)$ depends on the initialization and frequency. For ReLU activation function, 
        \begin{equation*}
            (\gamma(\vxi))^{2}=\Exp_{a(0), r(0)}\left[\frac{r(0)^{3}}{16 \pi^{4}\norm{\vxi}^{d+3}}+\frac{a(0)^{2} r(0)}{4 \pi^{2}\norm{\vxi}^{d+1}}\right],
        \end{equation*}
        where $r(0)=\abs{\vw(0)}$ and the two-layer NN parameters at initial $a(0)$ and $\vw(0)$ are random variables with certain given distribution. 
        The reason why there are $1/\norm{\vxi}^{d+1}$ and $1/\norm{\vxi}^{d+3}$ is explained with example $d=1$ as follows. In the gradient flow, the gradient of $h(x,\vtheta)$ w.r.t. $a_j$ leaves a ReLU function, which decays in $1/|\xi|^2$, and the gradient of $h(x,\vtheta)$ w.r.t. $w_j$ leaves a Heaviside function, which decays in $1/|\xi|$. Due to the chain rule, when we considering the evolution of the function, there will be square of gradients, thus, there are $1/\norm{\vxi}^{2}$ and $1/\norm{\vxi}^{4}$.
    
        The dynamics in equation (\ref{eq:ReLUnnFP}) shows the low frequency components converges faster than high frequency components. The RHS is only affected by the training data due to the sample distribution, that is, this dynamics stops evolution only when the error on training data is zero. However, the LHS contains the Fourier transform of the whole fitting function. Therefore, the fitting function beyond the training data can be inferred. The number of functions that satisfy the constraints on training data is infinite. It can be proved that the faster convergence of low-frequency components leads to that the model prefers to select a function with large low-frequency coefficients. \cite{zhang_explicitizing_2019,Luotao2020LFP} rigorously deduced the solution of the LFP model (\ref{eq:ReLUnnFP}) is equivalent to that of the following optimization problem in a proper hypothesis space $F_\gamma$,
        \begin{equation*}
            \min_{h-h_{\mathrm{ini}}\in F_{\gamma}}\int_{\sR^d}(\gamma(\vxi))^{-2}\abs{\fF[h](\vxi)-\fF[h_\mathrm{ini}](\vxi)}^{2}\diff{\vxi},\label{eq: minFPnorm-1}
        \end{equation*}
        subject to constraints $h(\vx_{i})=y_{i}$ for $i=1,\ldots,n$. Here, $h_{\mathrm{ini}} = h(\vx, \vtheta(0))$ is the original output of DNN. The weight $(\gamma(\vxi))^{-2}$ grows as the frequency $\vxi$ increases, which means that a large penalty is imposed on the high frequency part of $h(\vx)-h_{\mathrm{ini}}(\vx)$.
        Based on such analysis, $h(x)-h_{\mathrm{ini}}(x)$ is a low-frequency function (which is a function whose Fourier transform has large value only near the origin). But if $h_{\mathrm{ini}}(x)$ is not 0, $h(x)$ may be not a low-frequency function. In most cases, the target function in the real world is a low-frequency function. Thus, the low-frequency function is usually more likely to have good generalization in real data set. Therefore, a random non-zero initial output of DNN leads to a specific type of generalization error. To eliminate this error, we use DNNs with an antisymmetrical initialization (ASI) trick ~\citep{zhang2020type}, which guarantees $h_{\mathrm{ini}}(\vx)=0$. Then the final output $h(\vx)$ is dominated by low frequency, and the DNN model possesses a good generalization.

    \subsection{Fourier-domain Variational Formulation}
        
        Inspired by the variational formulation of LFP model, we propose a new continuum model for the supervised learning, which includes DNNs with gradient flow learning. 
        In the minimization problem proposed above, for large $\norm{\vxi}$, the main component of the weight term is $\norm{\vxi}^{d+1}$. Therefore, after ignoring the constant and low-order term, we can use $\langle\vxi\rangle^{d+1}$ to replace this weight item, where $\langle\vxi\rangle=(1+\norm{\vxi}^2)^{\frac{1}{2}}$ is the ``Japanese bracket'' of $\vxi$. We use this form to connect our variational problem with Sobolev space, which is convenient for us to carry out theoretical analysis.
        
        Further, we can replace the index  $d+1$ by a more general constant $\alpha$ to be determined later. Thus we obtain the following variational problem:
        \begin{align}
            & \min_{h\in \fH} Q_{\alpha}[h] =\int_{\sR^d}\langle\vxi\rangle^\alpha\Abs{\fF[h](\vxi)}^{2}\diff{\vxi},\label{mini-prob} \\
            & \mathrm{s.t.}\quad h(\vx_i)=y_i,\quad i=1,\cdots,n,
        \end{align}
        where $\fH=\{h(x)|\int_{\sR^d}\langle\vxi\rangle^\alpha\Abs{\fF[h](\vxi)}^{2}\diff{\vxi}<\infty\}$. According to the equivalent theorem in \cite{Luotao2020LFP}, $-\alpha$ is the decaying rate w.r.t. frequency in the gradient flow dynamics in (\ref{mini-prob}). In this work, we study how the property of the solution in the variational problem depends on $\alpha$.

        The solution of the variational problem is assumed to be not less-regular than $L^2$ functions. But in the spatial domain, the evaluation on $n$ known data points is meaningless in the sence of $L^2$ function since an $L^2$ function is defined upto a Lebesgue zero measure set.
        To overcome this, we consider the problem in the frequency domain and define a linear operator $\fP_{\vX}:L^1(\sR^d)\cap L^2(\sR^d)\to\sR^n$ for the given sample set $\vX$ to transform the original constraints into the ones in the Fourier domain: $\fP_{\vX}\phi^*=\vY$. More precisely, we define for $\phi\in L^1(\sR^d)\cap L^2(\sR^d)$
        \begin{equation}
            \fP_{\vX}\phi:=\left(\int_{\sR^d}\phi(\vxi)\E^{2\pi\I\vxi\cdot \vx_{1}}\diff{\vxi},\cdots,
            \int_{\sR^d}\phi(\vxi)\E^{2\pi\I\vxi\cdot \vx_{n}}\diff{\vxi}\right)^\T.
        \end{equation}
        \noindent The admissible function class reads as 
        \begin{equation*}
            \fA_{\vX,\vY}=\{\phi\in L^1(\sR^d)\cap L^2(\sR^d)\mid\fP_{\vX}\phi=\vY\}.
        \end{equation*}
        
        Notice that
        $\norm{\fF^{-1}[\phi]}_{H^{\frac{\alpha}{2}}}=\left(\int_{\sR^d}\langle\vxi\rangle^\alpha\abs{\phi(\vxi)}^{2}\diff{\vxi}\right)^{\frac{1}{2}}$ is a Sobolev norm, which characterizes the regularity of the final output function $h(\vx)=\fF^{-1}[{\phi}](\vx)$. The larger the exponent $\alpha$ is, the better the regularity becomes. For example, when $d=1$ and $\alpha=2$, by Parseval's theorem,
        \begin{equation*}
            \norm{u}_{H^{1}}^2=\int_{\sR}(1+\abs{\xi}^2)\abs{\fF[u](\xi)}^{2}\diff{\xi}=\int_{\sR}u^2+\frac{1}{4\pi^2}\abs{\nabla u}^2\diff x.
        \end{equation*}
        Accordingly, the Fourier-domain variational problem reads as a standard variational problem in spatial domain. This is true for any quadratic Fourier-domain variational problem, but of course our Fourier-domain variational formulation is not necessarily being quadratic. The details for general cases (non-quadratic ones) are left to future work. For the quadratic setting with exponent $\alpha$, i.e., Problem~\eqref{mini-prob}, it is roughly equivalent to the following spatial-domain variational problem:
        \begin{equation*}
            \min \int_{\sR^d}(u^2+\abs{\nabla ^{\frac{\alpha}{2}}u}^2)\diff x.
        \end{equation*}
        This is clear for integer $\alpha/2$, while fractional derivatives are required for non-integer $\alpha/2$. 
        
        Back to our problem, after the above transformation, our goal is transformed into studying the following Fourier-domain variational problem,
        \begin{prob}\label{prob..VariationalPointCloud}
        	Find a minimizer $\phi^*$ in $\fA_{\vX,\vY}$ such that
        	\begin{equation}
            	 \phi^*\in\arg\min_{\phi\in \fA_{\vX,\vY}} \norm{\fF^{-1}[\phi]}_{H^{\frac{\alpha}{2}}}^2.
        	\end{equation}
        \end{prob}
        
        We should point out that this variational problem is a paradigm for supervised learning problems, and a two-layer infinitely wide neural network corresponds to the special case of taking $\alpha=d+1$. And not all $\alpha$ can be obtained by the neural network with commonly-used activation functions. However, by analogy, we believe that the infinitely wide multi-layer neural network in NTK regime also belongs to the paradigm we proposed and corresponds the case $\alpha>d$.
        
        We remark that the operator $\fP_{\vX}$ is the inverse Fourier transform with evaluations on sample points $\vX$. Actually, the linear operator $\fP_{\vX}$ projects a function defined on $\sR^d$ to a function defined on $0$-dimensional manifold $\vX$. Just like the (linear) trace operator $T$ in a Sobolev space projects a function defined on $d$-dimensional manifold into a function defined on $(d-1)$-dimensional boundary manifold. Note that the only function space over the $0$-dimensional manifold $\vX$ is the $n$-dimensional vector space $\sR^n$, where $n$ is the number of data points, while any Sobolev (or Besov) space over $d$-dimensional manifold ($d\geq 1$) is an infinite dimensional vector space.

\section{The Critical Decaying Rate}\label{section4}
    
    In this section, we consider a critical exponent for $\alpha$, which leads to the existence/non-existence dichotomy to Problem~\ref{prob..VariationalPointCloud}. We first prove that there is no solution to the Problem~\ref{prob..VariationalPointCloud} in subcritical case $\alpha < d$, and for $\alpha>d$ the optimal function is a continuous and nontrivial solution (See proof in Appendix.). Therefore, we conclude that to obtain a non-trivial interpolation among training data for supervised learning, such as DNN fitting, the 
    decaying rate of high-frequency information can not be too fast, i.e., there exists a upper limit of the decaying rate w.r.t. frequency. 
    \subsection{Subcritical Case: $\alpha<d$}  
        In order to prove the nonexistence of the solution to the Problem~\ref{prob..VariationalPointCloud} in $\alpha<d$ case, at first we need to find a class of functions that make the norm tend to zero. 
        Let $\psi_{\sigma}(\vxi)=(2\pi)^{\frac{d}{2}}\sigma^d \E^{-2\pi^2\sigma^2\norm{\vxi}^2}$ , then by direct calculation, we have $\fF^{-1}[\psi_{\sigma}](\vx)=\E^{-\frac{\norm{\vx}^2}{2\sigma^2}}$. For $\alpha<d$ the following proposition shows that the norm $\norm{\fF^{-1}[\psi_\sigma]}_{H^{\frac{\alpha}{2}}}^2$ can be sufficiently small as $\sigma\rightarrow 0$.

        \begin{prop}[critical exponent]\label{prop..CriticalExponent}
        	For any input dimension $d$, we have
        	\begin{equation}
        	\lim_{\sigma\to0}\norm{\fF^{-1}[\psi_\sigma]}_{H^{\frac{\alpha}{2}}}^2
        	=\begin{cases}
        	0,      & \alpha<d, \\
        	C_d,      & \alpha=d, \\
        	\infty, & \alpha>d.
        	\end{cases}
        	\end{equation}
        	Here the constant $C_d=\frac{1}{2}(d-1)!(2\pi)^{-d}\frac{2\pi^{d/2}}{\Gamma\left(d/2\right)}$ only depends on the dimension $d$.
        \end{prop}
        
        \begin{rmk}
        	The function $\fF^{-1}[\psi]$ can be any function in the Schwartz space, not necessarily Gaussian. Proposition~\ref{prop..CriticalExponent} still holds with (possibly) different $C_d$.
        \end{rmk}
        For every small $\sigma$, we can use $n$ rapidly decreasing functions $\fF^{-1}[\psi_{\sigma}](\vx-\vx_{i})$ to construct the solution $\fF^{-1}[\phi_{\sigma}](\vx)$ of the supervised learning problem. However, according to Proposition~\ref{prop..CriticalExponent}, when the parameter $\sigma$ tends to 0, the limit is the zero function in the sense of $L^2(\sR^d)$. Therefore we have the following theorem:
        \begin{thm}[non-existence]\label{thm..alpha<d} 
        	Suppose that $\vY\neq\vzero$. For $\alpha<d$, there is no function $\phi^*\in \fA_{\vX,\vY}$ satisfying
        	\begin{equation*}
            	 \phi^*\in\arg\min_{\phi\in \fA_{\vX,\vY}}\norm{\fF^{-1}[\phi]}_{H^{\frac{\alpha}{2}}}^2.
        	\end{equation*}
        	In other words, there is no solution to the Problem~\ref{prob..VariationalPointCloud}. 
        \end{thm}
    
    
    \subsection{Supercritical Case: $\alpha>d$}\label{Supercritical case}
        We then provide a theorem to establish the existence of the minimizer for Problem \ref{prob..VariationalPointCloud} in the case of $\alpha>d$. 
        
        \begin{thm}[existence]\label{thm..alpha>d} 
        	For $\alpha>d$, there exists $\phi^*\in \fA_{\vX,\vY}$ satisfying
        	\begin{equation*}
            	 \phi^*\in\arg\min_{\phi\in \fA_{\vX,\vY}}\norm{\fF^{-1}[\phi]}_{H^{\frac{\alpha}{2}}}^2.
        	\end{equation*}
        	In other words, there exists a solution to the Problem \ref{prob..VariationalPointCloud}.
        \end{thm}
        \begin{rmk}
            Note that, according to the Sobolev embedding theorem~\citep{adams2003sobolev,1999Partial}, the minimizer in Theorem \ref{thm..alpha>d} has smoothness index no less than $[\frac{\alpha-d}{2}]$.
        \end{rmk}


\section{Numerical Results}\label{section5}
    In this section, we illustrate our results by solving Fourier-domain variational problems numerically. We use uniform mesh in frequency domain with mesh size $\Delta\xi$ and band limit $M\Delta\xi$. In this discrete setting, the considered space becomes $\sR^{(2M)^d}$. We emphasize that the numerical solution with this setup always exists even for the subcritical case which corresponds to the non-existence theorem. However, as we will show later, the numerical solution is trivial in nature when $\alpha<d$.
    
    \subsection{Special Case: One Data Point in One Dimension}\label{special_case}
        To simplify the problem, we start with a single point $X=0\in\sZ$ with the label $Y=2$. Denote $\phi_j = \phi(\xi_j)$ for $j\in\sZ$. We also assume that the function $\phi$ is an even function. Then according to the definition of $\fP_{\vX}$, we have the following problem:
        \begin{exam}[Problem~\ref{prob..VariationalPointCloud} with a particular discretization]  
        	\begin{align}
            	& \min_{\phi\in\sR^M} \sum_{j=1}^M(1+{j}^2\Delta\xi^2)^{\frac{\alpha}{2}}\Abs{\phi_j}^{2}, \\
            	& \mathrm{s.t.}\quad \sum_{j=1}^M\phi_j\Delta\xi = 1,
        	\end{align}
        \end{exam}

        If we denote $\vphi = {(\phi_1, \phi_2, \ldots, \phi_M)}^{\T}$, $b = \frac{1}{\Delta\xi}$, $\mA = (1, 1, \ldots, 1)\in\sR^M$ and $\mGamma \in \sR^M$ where $\mGamma_{ii}=\sqrt{\lambda}(1+i^2\Delta\xi^2)^{\frac{\alpha}{4}},\ i=1,2,\cdots,M.$
        
        In fact this is a standard Tikhonov regularization~\citep{1977Solutions}, also known as ridge regression problem with the Lagrange multiplier $\lambda$. The corresponding ridge regression problem is,
        \begin{equation}
        	\min_{\vphi}{\norm{\mA\vphi - b}_2^2 + \norm{\mGamma\vphi}_2^2},
        \end{equation}
        This problem admits an explicit and unique solution~\citep{1977Solutions},
        \begin{equation}\label{ridge_solution}
        	\vphi = {(\mA^{\T}\mA + \mGamma^{\T}\mGamma)}^{-1}\mA^{\T} b.
        \end{equation}
        
        By using the inverse discrete Fourier transform, we obtain the final output function in $x$ space:
        \begin{equation}\label{h(x)}
        	h(x) = \frac{2}{(Z^2 + \lambda)} \sum_{j=1}^{M} (1+j^2\Delta\xi^2)^{-\frac{\alpha}{2}} \cos(2\pi j x).
        \end{equation}
        
\begin{figure} 
        	\centering
            \includegraphics[width=0.47\textwidth]{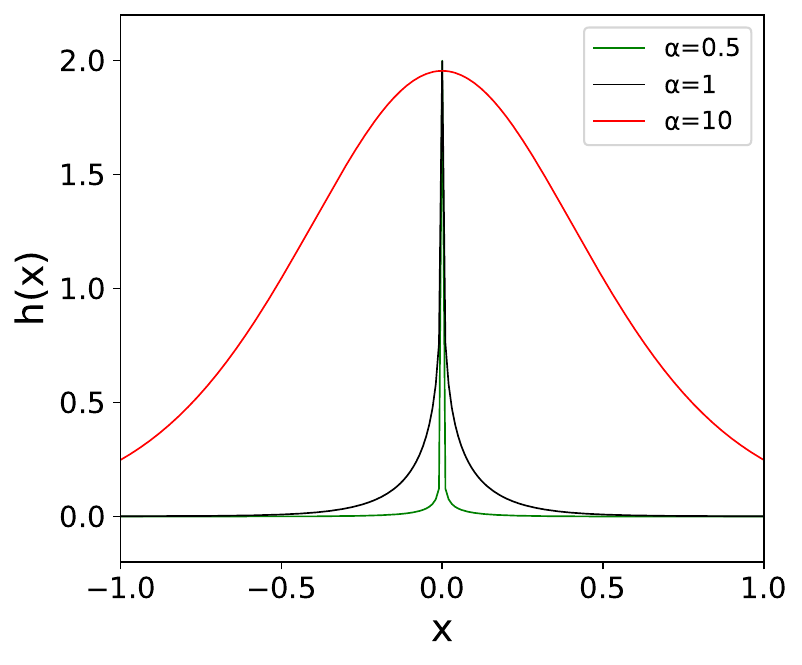}
        	\caption{Fitting the function $h(x)$ shown in equation~\eqref{h(x)} with different exponent $\alpha$'s. Here we take $M=10^6$, $\Delta\xi=0.01$, $\lambda=1$ and different $\alpha$ and observe that $h(x)$ is not an trivial function in $\alpha>d$ case and degenerates to a trivial function in $\alpha<d$ case.}
        	\label{fig:hx_is_nontrivial}
        \end{figure}

        Fig.~\ref{fig:hx_is_nontrivial}  shows that for this special case with a large $M$, $h(x)$ is not an trivial function in $\alpha>d$ case and degenerates to a trivial function in $\alpha<d$ case.
        
    \subsection{General Case: $n$ Points in $d$ Dimension}\label{case2}
        Assume that we have $n$ data points $\vx_1,\vx_2,\ldots,\vx_n\in \sR^d$ and each data point has $d$ components which means $\vx_i=\left(x_{i1},x_{i2},\ldots,x_{id}\right)^\T$,
        and denote the corresponding label as $\left(y_1,y_2,\ldots,y_n\right)^\T$. For the sake of simplicity, we denote the vector $(j_1,j_2,\cdots,j_d)^\T$ by $\vJ_{j_1\ldots j_d}$. Then our problem becomes
        \begin{exam}[Problem \ref{prob..VariationalPointCloud} with general discretization]
        	\begin{align}
        	& \min_{\phi\in \sR^{(2M)^d}} \sum_{j_1,\ldots,j_d=-M}^M(1+\norm{\vJ_{j_1\ldots j_d}}^2\Delta\xi^2)^{\frac{\alpha}{2}}\Abs{\phi_{j_1\ldots j_d}}^{2}, \\
        	& \mathrm{s.t.}\quad \sum_{j_1,\ldots,j_d=-M}^M\phi_{j_1\ldots j_d}\E^{2\pi\I\Delta\xi\vJ_{j_1\ldots j_d}^\T\vx_k}=y_k, \ \ k=1,2,\ldots,d
        	\end{align}
        \end{exam}
        
        The calculation of this example can be completed by the method analogous to the one used in Example~\ref{special_case}.
        
\begin{figure}
            \centering
        	\subfigure[2 points in 1 dimension]{
        	    \includegraphics[width=0.47\textwidth]{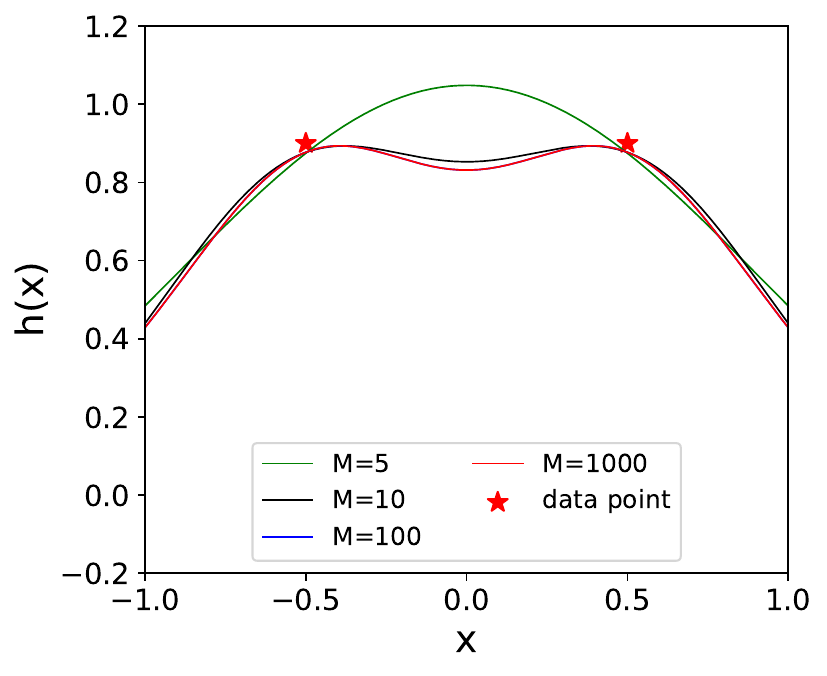}}
            \subfigure[20 points in 2 dimension]
                {\includegraphics[width=0.47\textwidth]{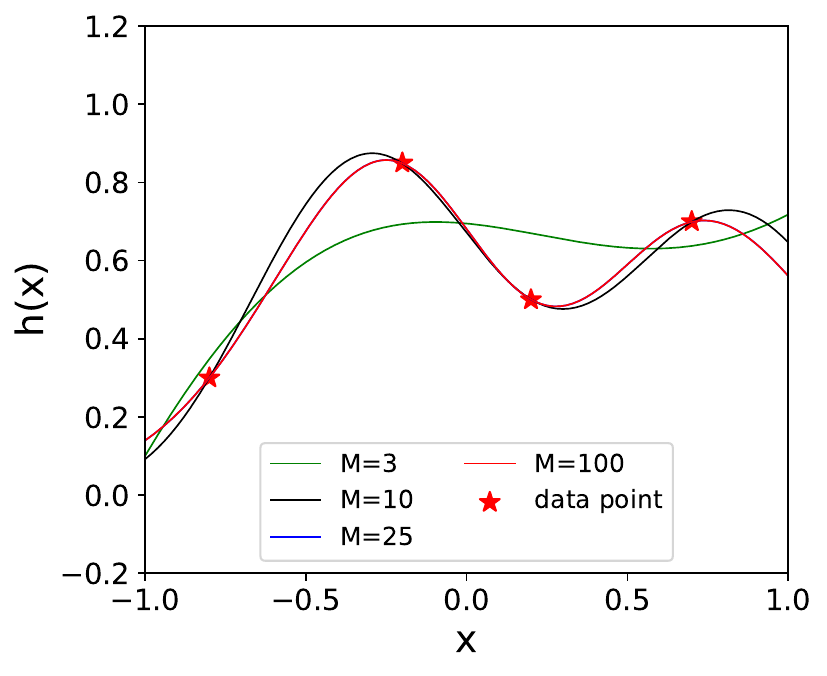}}
        	\caption{Fitting data points in different dimensions with different band limit $M$. We use a proper $\alpha$ ($\alpha>d$) and observe that even for a large $M$, the function $h(x)$ does not degenerate to a trivial function. Note that the blue curve and the red one overlap with each. Here the trivial function represents a function whose value decays rapidly to zero except for the given training points.}
        	\label{fig:diff_M}
        \end{figure}

\begin{figure}
        	\centering
        	\subfigure[2 points in 1 dimension]
        	{
        		\includegraphics[width=0.47\textwidth]{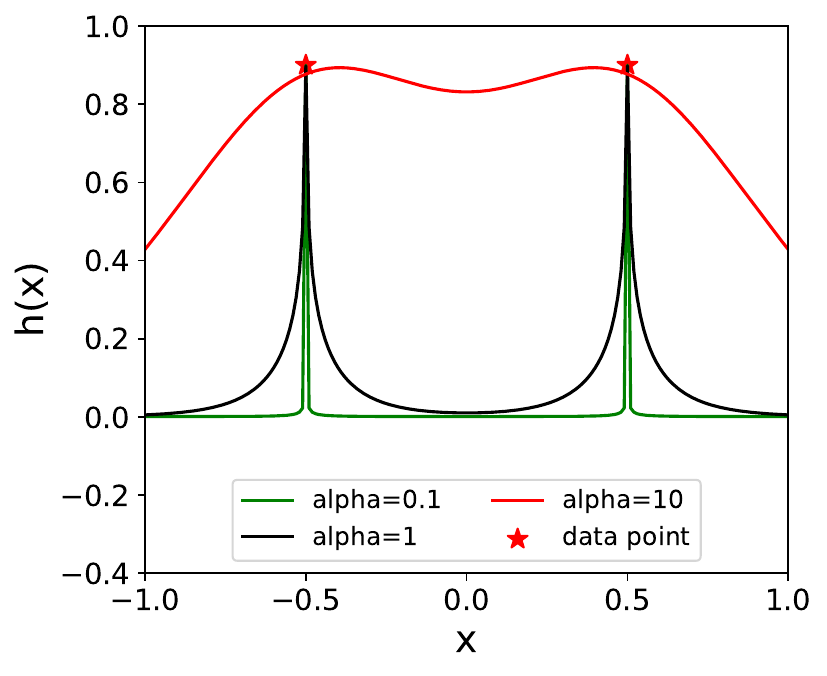}
        	}
        	\subfigure[20 points in 2 dimension]{
        		\includegraphics[width=0.47\textwidth]{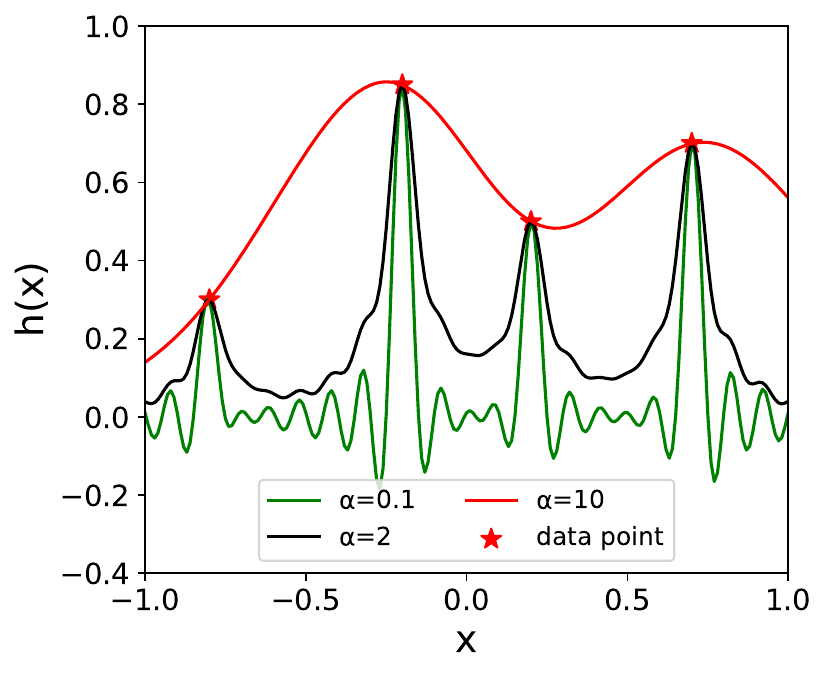}
        	}
        	\caption{Fitting data points in different dimensions with different exponent $\alpha$'s. We observe that with a proper $M$, the function $h(x)$ is not a trivial function for $\alpha>d$ case and degenerates to a trivial function for $\alpha<d$ case.}
        	\label{fig:diff_alpha}
        \end{figure}

         In Fig.\ref{fig:diff_M}, we set $\alpha=10$ in both cases to ensure $\alpha>d$ and change the band limit $M$. We observe that as $M$ increases, the fitting curve converges to a non-trivial curve. In Fig.\ref{fig:diff_alpha}, we set $M=1000$ in 1-dimensional case and $M=100$ in 2-dimensional case. By changing exponent $\alpha$, we can see in all cases, the fitting curves are non-trivial when $\alpha>d$, but degenerate when $\alpha<d$.

        %
        %

\section{Conclusion \label{sec:conclusion}}
    To understand the limit of the frequency bias in DNNs, we propose a Fourier-domain variational formulation based on the LFP model and establish the sufficient and necessary conditions for the well-posedness of the Fourier-domain variational problem, followed by numerical demonstration. Our work suggests that there is a upper limit of the decaying rate w.r.t. frequency, i.e., high frequency cannot converge too fast, in order to obtain a nontrivial solution in LFP model, thus,  pointing out the intrinsic high-frequency curse. For two-layer infinite-width neural networks, existing works have shown their solutions are equivalent to the solutions of particular Fourier-domain variational problems \citep{Luotao2020LFP}. However, for general non-linear DNNs, this equivalence is only qualitative. 
    
    In addition, our Fourier-domain variational formulation provides a novel viewpoint for modelling machine learning problem, that is, imposing more constraints, e.g., higher regularity, on the model rather than the data (always isolated points in practice) can give us the well-posedness as dimension of the problem increases. This is different from the modelling in physics and traditional point cloud problems, in which the model is independent of dimension in general. Our work suggests a potential approach of algorithm design by considering a dimension-dependent model for data modelling.

\acks{This work is sponsored by the National Key R\&D Program of China  Grant No. 2019YFA0709503 (Z. X.) and No. 2020YFA0712000 (Z. M.), the Shanghai Sailing Program (Z. X.), the Natural Science Foundation of Shanghai Grant No. 20ZR1429000  (Z. X.), the National Natural Science Foundation of China Grant No. 62002221 (Z. X.), the National Natural Science Foundation of China Grant No. 12101401 (T. L.), the National Natural Science Foundation of China Grant No. 12101402 (Y. Z.), Shanghai Municipal of Science and Technology Project Grant No. 20JC1419500 (Y. Z.), the Lingang Laboratory Grant No.LG-QS-202202-08 (Y.Z.), the National Natural Science Foundation of China Grant No. 12031013 (Z. M.), Shanghai Municipal of Science and Technology Major Project No. 2021SHZDZX0102, and the HPC of School of Mathematical Sciences and the Student Innovation Center at Shanghai Jiao Tong University.}

\bibliography{DLRef_1}

\appendix

\section{Lemma~\ref{lemma1}}

    \begin{lem}\label{lemma1}
    	Let the function $\psi_{\sigma}(\vxi)=(2\pi)^{\frac{d}{2}}\sigma^d \E^{-2\pi^2\sigma^2\norm{\vxi}^2}$, $\vxi\in\sR^d$. We have
    	\begin{equation}\label{lemmaeq}
    		\lim_{\sigma\to0}\int_{\sR^d}\norm{\vxi}^{\alpha}\abs{\psi_{\sigma}(\vxi)}^{2}\diff{\vxi}
    		=\begin{cases}
    			0,      & \alpha<d, \\
    			C_d,      & \alpha=d, \\
    			\infty, & \alpha>d.
    		\end{cases}
    	\end{equation}
    	Here the constant $C_d=\frac{1}{2}(d-1)!(2\pi)^{-d}\frac{2\pi^{d/2}}{\Gamma\left(d/2\right)}$ only depends on the dimension $d$.
    \end{lem}
    
    \begin{proof} 
    	In fact, 
    	\begin{align*}
    		\lim\limits_{\sigma\rightarrow 0}  \int_{\sR^d}\norm{\vxi}^\alpha|\psi_\sigma(\vxi)|^2\diff\vxi
    		&=\lim\limits_{\sigma\rightarrow 0}  \int_{\sR^d}\norm{\vxi}^\alpha(2\pi)^d\sigma^{2d}\E^{-4\pi^2\sigma^2\norm{\vxi}^2}\diff\vxi\\
    		&=\lim\limits_{\sigma\rightarrow 0}  (2\pi)^d  \sigma^{d-\alpha}  \int_{\sR^d}  \norm{\sigma\vxi}^\alpha  \E^{-4\pi^2\norm{\sigma\vxi}^2}\diff{(\sigma\vxi)}\\
    		&=\lim\limits_{\sigma\rightarrow 0}  (2\pi)^d  \sigma^{d-\alpha}  \int_0^\infty r^{\alpha+d-1}  \E^{-4\pi^2r^2}\diff r\cdot \omega_d,
    	\end{align*}
    	where $\omega_d=\frac{2\pi^{\frac{d}{2}}}{\Gamma\left(\frac{d}{2}\right)}$ is the surface area of a unit $(d-1)$-sphere.
    	
    	\noindent Notice that 
    	\begin{align*}
    		\int_0^\infty r^{\alpha+d-1}  \E^{-4\pi^2r^2}\diff r&= 
    		\int_0^1 r^{\alpha+d-1}  \E^{-4\pi^2r^2}\diff r + \int_1^\infty r^{\alpha+d-1}  \E^{-4\pi^2r^2}\diff r\\
    		&\le \int_0^\infty \E^{-4\pi^2r^2}\diff r + \int_0^\infty r^{[\alpha]+d}  \E^{-4\pi^2r^2}\diff r\\
    		&=\frac{1}{8\pi^{\frac{3}{2}}} + \int_0^\infty r^{[\alpha]+d}  \E^{-4\pi^2r^2}\diff r
    	\end{align*}
    	and
    	\begin{equation*}
    		\int_0^\infty r^{[\alpha]+d}  \E^{-4\pi^2r^2}\diff r=
    		\begin{cases}
    			\frac{1}{2}\left(\frac{[\alpha]+d-1}{2}\right)!(2\pi)^{-([\alpha]+d+1)}, & [\alpha]+d\mathrm{\ is\ odd},\\
    			\frac{\sqrt{\pi}}{2}(2\pi)^{-([\alpha]+d+1)}(\frac{1}{2})^{\frac{[\alpha]+d}{2}}([\alpha]+d-1)!!, & [\alpha]+d\mathrm{\ is\ even}.
    		\end{cases}
    	\end{equation*}
    	Therefore, in both cases, the integral $\int_0^\infty r^{\alpha+d-1}  \E^{-4\pi^2r^2}\diff r$ is finite. Then we have 
    	\begin{align*}
    		\lim\limits_{\sigma\rightarrow 0}  \int_{\sR^d}\norm{\vxi}^\alpha|\psi_\sigma(\vxi)|^2\diff\vxi&=\lim\limits_{\sigma\rightarrow 0}  (2\pi)^d  \sigma^{d-\alpha}  \int_0^\infty r^{\alpha+d-1}  \E^{-4\pi^2r^2}\diff r\cdot \omega_d\\
    		&=\begin{cases}
    			0,      & \alpha<d, \\
    			\infty, & \alpha>d.
    		\end{cases}
    	\end{align*}
    	
    	\noindent When $\alpha=d$, it follows that
    	\begin{equation*}
    		\int_0^\infty r^{\alpha+d-1}  \E^{-4\pi^2r^2}\diff r
    		=\frac{1}{2}(2\pi)^{-2d}(d-1)!.
    	\end{equation*}
    	
    	\noindent Therefore
    	\begin{equation*}
    		\lim\limits_{\sigma\rightarrow 0}  \int_{\sR^d}\norm{\vxi}^\alpha|\psi_\sigma(\xi)|^2\diff\xi
    		=\frac{1}{2}(d-1)!(2\pi)^{-d}\frac{2\pi^{\frac{d}{2}}}{\Gamma\left(\frac{d}{2}\right)},
    	\end{equation*}
    	which completes the proof.
    \end{proof}

\section{Proof of Proposition~\ref{prop..CriticalExponent}}
    
    \begin{proof}  
    	Similar to the proof of Lemma \ref{lemma1}, we have
    	\begin{align*}
    		\lim\limits_{\sigma\rightarrow 0}  \norm{\fF^{-1}[\psi_\sigma]}_{H^{\frac{\alpha}{2}}}^2
    		&=\lim\limits_{\sigma\rightarrow 0}  (2\pi)^d  \sigma^{d-\alpha}  \int_{\sR^d}  (\sigma^2+\norm{\sigma\vxi}^2)^{\frac{\alpha}{2}}  \E^{-4\pi^2\norm{\sigma\vxi}^2}\diff(\sigma\vxi)\\
    		&=\lim\limits_{\sigma\rightarrow 0}  (2\pi)^d  \sigma^{d-\alpha}  \int_0^\infty  r^{d-1}(\sigma^2+r^2)^{\frac{\alpha}{2}}  \E^{-4\pi^2r^2}\diff r\cdot \omega_d.\\
    	\end{align*}
    	For $\sigma<1$, the following integrals are bounded from below and above, respectively:
    	\begin{equation*}
    		\int_0^\infty  r^{d-1}(\sigma^2+r^2)^{\frac{\alpha}{2}}  \E^{-4\pi^2r^2}\diff r \ge \int_0^\infty  r^{\alpha+d-1}  \E^{-4\pi^2r^2}\diff r = C_1 >0,
    	\end{equation*}
    	and
    	\begin{align*} 
    		\int_0^\infty  r^{d-1}(\sigma^2+r^2)^{\frac{\alpha}{2}}  \E^{-4\pi^2r^2}\diff r &\le
    		\int_0^1 r^{d-1}(1+r^2)^{\frac{\alpha}{2}}  \E^{-4\pi^2r^2}\diff r + 
    		\int_1^\infty r^{d-1}((2r)^2)^{\frac{\alpha}{2}}  \E^{-4\pi^2r^2}\diff r \\&\le
    		\int_0^1 r^{d-1}(1+r^2)^{\frac{\alpha}{2}}  \E^{-4\pi^2r^2}\diff r +
    		2^\alpha\int_0^\infty r^{\alpha+d-1}  \E^{-4\pi^2r^2}\diff r\\&= C_2 < \infty,
    	\end{align*}
    	where $C_1=\int_0^\infty  r^{\alpha+d-1}  \E^{-4\pi^2r^2}\diff r$ and $C_2=\int_0^1 r^{d-1}(1+r^2)^{\frac{\alpha}{2}}  \E^{-4\pi^2r^2}\diff r +
    	2^\alpha\int_0^\infty r^{\alpha+d-1}  \E^{-4\pi^2r^2}\diff r$. Therefore, we obtain the results for the subcritical ($\alpha<d$) and supercritical ($\alpha>d$) cases
    	\begin{align*}
    		\lim\limits_{\sigma\rightarrow 0}  \norm{\fF^{-1}[\psi_\sigma]}_{H^{\frac{\alpha}{2}}}^2
    		&=\lim\limits_{\sigma\rightarrow 0}  (2\pi)^d  \sigma^{d-\alpha}  \int_0^\infty  r^{d-1}(\sigma^2+r^2)^{\frac{\alpha}{2}}  \E^{-4\pi^2r^2}\diff r\cdot \omega_d\\
    		&=\begin{cases}
    			0,      & \alpha<d, \\
    			\infty, & \alpha>d.
    		\end{cases}
    	\end{align*}
    	
    	\noindent For the critical case $\alpha=d$, we have
    	\begin{align*}
    		&\lim\limits_{\sigma\rightarrow 0}  \norm{\fF^{-1}[\psi_\sigma]}_{H^{\frac{\alpha}{2}}}^2\\
    		&=\lim\limits_{\sigma\rightarrow 0}  (2\pi)^d \int_0^\infty  r^{d-1}(\sigma^2+r^2)^{\frac{\alpha}{2}}  \E^{-4\pi^2r^2}\diff r\cdot \omega_d\\
    		&=\lim\limits_{\sigma\rightarrow 0}  (2\pi)^d \int_0^\infty  r^{2d-1}  \E^{-4\pi^2r^2}\diff r\cdot \omega_d + \lim\limits_{\sigma\rightarrow 0}\left[\frac{\alpha}{2}(2\pi)^d \sigma^2\int_0^\infty r^{2d-3}\E^{-4\pi^2r^2}\diff r\cdot \omega_d+o(\sigma^2)\right]\\
    		&=\lim\limits_{\sigma\rightarrow 0}  (2\pi)^d \int_0^\infty  r^{2d-1}  \E^{-4\pi^2r^2}\diff r\cdot \omega_d\\
    		&=\frac{1}{2}(d-1)!(2\pi)^{-d}\frac{2\pi^{\frac{d}{2}}}{\Gamma\left(\frac{d}{2}\right)}.
    	\end{align*}
    	\noindent Therefore the proposition holds.
    \end{proof}

\section{Proof of Theorem~\ref{thm..alpha<d}}
    
    \begin{proof} 
    	Given $\vX=(\vx_{1},\ldots,\vx_{n})^\T$ and $\vY=(y_{1},\ldots,y_{n})^\T$, let $\mA=\left(\exp(-\frac{\norm{\vx_j-\vx_i}^2}{2\sigma^2})\right)_{n\times n}$ be an $n\times n$ matrix. For sufficiently small $\sigma$, the matrix $\mA$ is diagonally dominant, and hence invertible. So the linear system $\mA\vg^{(\sigma)}=\vY$ has a solution $\vg^{(\sigma)}=\left(g^{(\sigma)}_1,g^{(\sigma)}_2,\cdots,g^{(\sigma)}_n\right)^\T$. Let
    	\begin{equation*}
    		\phi_{\sigma}(\vxi)
    		=\sum_{i}g^{(\sigma)}_{i}\E^{-2\pi\I\vxi^\T\vx_i}\psi_{\sigma}(\vxi),
    	\end{equation*}
    	\noindent where $\psi_{\sigma}(\vxi)=(2\pi)^{\frac{d}{2}}\sigma^d\E^{-2\pi^2\sigma^2\norm{\vxi}^2}$ satisfying $\fF^{-1}[\psi_{\sigma}](\vx)=\E^{-\frac{\norm{\vx}^2}{2\sigma^2}}$. Thus
    	\begin{equation*}
    		\fF^{-1}[\phi_{\sigma}](\vx)
    		=\sum_{i}g^{(\sigma)}_{i}\fF^{-1}[\psi_{\sigma}](\vx-\vx_{i})
    		=\sum_{i}g^{(\sigma)}_{i}\E^{-\frac{\norm{\vx-\vx_{i}}^2}{2\sigma^2}}.
    	\end{equation*}
    	In particular, for all $i=1,2,\cdots,n$
    	\begin{equation*}
    		\fF^{-1}[\phi_{\sigma}](\vx_i)
    		=\sum_{j}g^{(\sigma)}_{j}\E^{-\frac{\norm{\vx_i-\vx_{j}}^2}{2\sigma^2}}=(\mA\vg^{(\sigma)})_i=y_i.
    	\end{equation*}
    	\noindent Therefore, $\phi_{\sigma}\in \fA_{\vX,\vY}$ for sufficiently small $\sigma>0$.
    	
    	According to the above discussion, we can construct a sequence $\{\phi_{\frac{1}{m}}\}_{m=M}^\infty\subset \fA_{\vX,\vY}$, where $M$ is a sufficiently large positive integer to make the matrix $\mA$ invertible. As Proposition~\ref{prop..CriticalExponent} shows,
    	\begin{equation*}
    		\lim_{m\to+\infty}\norm{\fF^{-1}[\phi_{\frac{1}{m}}]}_{H^{\frac{\alpha}{2}}}^2=0.
    	\end{equation*}
    	Now, suppose that there exists a solution to the Problem \ref{prob..VariationalPointCloud}, denoted as $\phi^*\in \fA_{\vX,\vY}$. By definition,
    	\begin{equation*}
    		\norm{\fF^{-1}[\phi^*]}_{H^{\frac{\alpha}{2}}}^2
    		\leq \min_{\phi\in \fA_{\vX,\vY}} \norm{\fF^{-1}[\phi]}_{H^{\frac{\alpha}{2}}}^2
    		\le\lim_{m\to+\infty}\norm{\fF^{-1}[\phi_{\frac{1}{m}}]}_{H^{\frac{\alpha}{2}}}^2=0.
    	\end{equation*}
    	Therefore, $\phi^*(\vxi)\equiv 0$ and $\fP_{\vX}\phi^*=\vzero$, which contradicts to the restrictive condition $\fP_{\vX}\phi^*=\vY$ for the situation that $\vY\neq \vzero$. The proof is completed.
    \end{proof}

\section{Proof of Theorem~\ref{thm..alpha>d}}
    
    \begin{proof}    
    	\noindent 1.
    	We introduce a distance for functions $\phi,\psi\in L^2(\sR^d)$:
    	\begin{equation*}
    		\dist(\phi,\psi)=\norm{\fF^{-1}[\phi]-\fF^{-1}[\psi]}_{H^{\frac{\alpha}{2}}}.
    	\end{equation*}
    	Under the topology induced by this distance, the closure of the admissible function class $\fA_{\vX,\vY}$ reads as
    	\begin{equation*}
    		\overline{\fA_{\vX,\vY}}:=\overline{\{\phi\in L^1(\sR^d)\cap L^2(\sR^d)\mid\fP_{\vX}\phi=\vY\}}^{\mathrm{dist}(\cdot,\cdot)}.
    	\end{equation*}
    	\noindent 2.
    	We will consider an auxiliary minimization problem: to find $\phi^*$ such that
    	\begin{equation}\label{eq..AuxiliaryMinProb}
    		\phi^*\in\arg\min_{\phi\in \overline{\fA_{\vX,\vY}}}\norm{\fF^{-1}[\phi]}_{H^{\frac{\alpha}{2}}}.
    	\end{equation}
    	Let $m:=\inf_{\phi\in \overline{\fA_{\vX,\vY}}}\norm{\fF^{-1}[\phi]}_{H^{\frac{\alpha}{2}}}$.
    	According to the proof of Proposition~\ref{prop..CriticalExponent} and Theorem~\ref{thm..alpha<d}, for a small enough $\sigma>0$, the inverse Fourier transform of function 
    	\begin{equation*}
    		\phi_{\sigma}(\vxi)
    		=\sum_{i}g^{(\sigma)}_{i}\E^{-2\pi\I\vxi^\T\vx_i}\psi_{\sigma}(\vxi)
    	\end{equation*}
    	has finite Sobolev norm $\norm{\fF^{-1}[\phi_\sigma]}_{H^{\frac{\alpha}{2}}}<\infty$, where $\psi_{\sigma}(\vxi)$ satisfies $\fF^{-1}[\psi_{\sigma}](\vx)=\E^{-\frac{\norm{\vx}^2}{2\sigma^2}}$, $\mA=\left(\exp(-\frac{\norm{\vx_j-\vx_i}^2}{2\sigma^2})\right)_{n\times n}$ and  $\vg^{(\sigma)}=\left(g^{(\sigma)}_1,g^{(\sigma)}_2,\cdots,g^{(\sigma)}_n\right)^\T=\mA^{-1}\vY$. Thus $m<+\infty$.
    	
    	\noindent 3.
    	Choose a minimizing sequence $\{\bar{\phi}_k\}_{k=1}^\infty\subset \overline{\fA_{\vX,\vY}}$ such that
    	\begin{equation*}
    		\lim_{k\rightarrow \infty} \norm{\fF^{-1}[\bar{\phi}_k]}_{H^{\frac{\alpha}{2}}} =m.
    	\end{equation*}
    	By definition of the closure, there exists a function $\phi_k\in\fA_{\vX,\vY}$ for each $k$ such that
    	\begin{equation*}
    		\norm{\fF^{-1}[\bar{\phi}_k]-\fF^{-1}[\phi_k]}_{H^{\frac{\alpha}{2}}}\leq \frac{1}{k}.
    	\end{equation*}
    	Therefore $\{\phi_k\}_{k=1}^\infty\subset \fA_{\vX,\vY}$ is also a minimizing sequence, i.e.,
    	\begin{equation*}
    		\lim_{k\rightarrow \infty} \norm{\fF^{-1}[\phi_k]}_{H^{\frac{\alpha}{2}}} =m.
    	\end{equation*}	
    	Then $\{\fF^{-1}[\phi_k]\}_{k=1}^\infty$ is bounded in the Sobolev space $H^{\frac{\alpha}{2}}(\sR^d)$. Hence there exist a weakly convergent subsequence $\{\fF^{-1}[\phi_{n_k}]\}_{k=1}^\infty$ and a function $\fF^{-1}[\phi^*]\in H^{\frac{\alpha}{2}}(\sR^d)$ such that
    	\begin{equation*} 
    		\fF^{-1}[\phi_{n_k}]\rightharpoonup \fF^{-1}[\phi^*] \quad\text{in } H^{\frac{\alpha}{2}}(\sR^d)\text{ as}\  k\rightarrow \infty.
    	\end{equation*}
    	Note that
    	\begin{equation*} 
    		m=\inf_{\phi\in \overline{\fA_{\vX,\vY}}}\norm{\fF^{-1}[\phi]}_{H^{\frac{\alpha}{2}}}\le 
    		\norm{\fF^{-1}[\phi^*]}_{H^{\frac{\alpha}{2}}}\le\liminf_{\phi_{n_k}}\norm{\fF^{-1}[\phi_{n_k}]}_{H^{\frac{\alpha}{2}}}=m,
    	\end{equation*}
    	where we have used the lower semi-continuity of the Sobolev norm of $H^{\frac{\alpha}{2}}(\sR^d)$ in the third inequality. Hence $\norm{\fF^{-1}[\phi^*]}_{H^{\frac{\alpha}{2}}}=m$.
    	
    	\noindent 4. We further establish the strong convergence that
    	$\fF^{-1}[\phi_{n_k}]-\fF^{-1}[\phi^*]\rightarrow 0$ in $H^{\frac{\alpha}{2}}(\sR^d)$ as $k\rightarrow \infty$.
    	
    	In fact, since $\fF^{-1}[\phi_{n_k}]\rightharpoonup \fF^{-1}[\phi^*] \ \text{in } H^{\frac{\alpha}{2}}(\sR^d)\text{ as}\ k\rightarrow \infty$ and $\lim_{k\rightarrow \infty}\norm{\fF^{-1}[\phi_{n_k}]}_{H^{\frac{\alpha}{2}}}=m=\norm{\fF^{-1}[\phi^*]}_{H^{\frac{\alpha}{2}}}$, we have
    	\begin{align*}
    		&\lim_{k\to \infty}\norm{\fF^{-1}[\phi_{n_k}]-\fF^{-1}[\phi^*]}_{H^{\frac{\alpha}{2}}}^2=\lim_{k\to \infty} \langle\fF^{-1}[\phi_{n_k}]-\fF^{-1}[\phi^*],\fF^{-1}[\phi_{n_k}]-\fF^{-1}[\phi^*]\rangle\\
    		&=\lim_{k\to \infty} \langle\fF^{-1}[\phi_{n_k}],\fF^{-1}[\phi_{n_k}]\rangle+ \langle\fF^{-1}[\phi^*],\fF^{-1}[\phi^*]\rangle- \langle\fF^{-1}[\phi_{n_k}],\fF^{-1}[\phi^*]\rangle- \langle\fF^{-1}[\phi^*],\fF^{-1}[\phi_{n_k}]\rangle\\
    		&=m^2+m^2- \lim_{k\to \infty}\left(\langle\fF^{-1}[\phi_{n_k}],\fF^{-1}[\phi^*]\rangle+ \langle\fF^{-1}[\phi^*],\fF^{-1}[\phi_{n_k}]\rangle\right)\\
    		&=m^2+m^2-\langle\fF^{-1}[\phi^*],\fF^{-1}[\phi^*]\rangle-\langle\fF^{-1}[\phi^*],\fF^{-1}[\phi^*]\rangle=0.
    	\end{align*}
    	Here $\langle\cdot,\cdot \rangle$ is the inner product of the Hilbert space $H^{\frac{\alpha}{2}}$.
    	
    	\noindent 5. 
    	We have $\phi^*\in L^1(\sR^d)$ because
    	\begin{align*}
    		\int_{\sR^d}\abs{\phi^*(\vxi)}\diff \vxi =\int_{\sR^d}\frac{\langle\vxi\rangle^{\frac{\alpha}{2}}\abs{\phi^*(\vxi)}}{\langle\vxi\rangle^{\frac{\alpha}{2}}}\diff \vxi
    		\le\norm{\fF^{-1}[\phi^*]}_{H^{\frac{\alpha}{2}}} \left(\int_{\sR^d}\frac{1}{\langle\vxi\rangle^\alpha}\diff\vxi\right)^{\frac{1}{2}}=Cm<+\infty,
    	\end{align*}
    	where $C:=\left(\int_{\sR^d}\frac{1}{\langle\vxi\rangle^\alpha}\diff\vxi\right)^{\frac{1}{2}}<+\infty$. Hence $\phi^*\in L^1(\sR^d)\cap L^2(\sR^d)$ and $\fP_{\vX}\phi^*$ is well-defined.
    	
    	\noindent 6. Recall that $\fP_{\vX}\phi_{n_k}=\vY$. We have
    	\begin{align*}
    		\Abs{\vY-\fP_{\vX}\phi^*}
    		&=\lim_{k\to+\infty}\Abs{\fP_{\vX}\phi_{n_k}-\fP_{\vX}\phi^*}\\
    		&=
    		\lim_{k\to+\infty}\Abs{\int_{\sR^d}(\phi_{n_k}-\phi^*)\E^{2\pi\I\vx\vxi}\diff\vxi}\\
    		&=\lim_{k\to+\infty}\Abs{\int_{\sR^d}\frac{\langle\vxi\rangle^{\frac{\alpha}{2}}(\phi_{n_k}-\phi^*)}{\langle\vxi\rangle^{\frac{\alpha}{2}}}\E^{2\pi\I\vx\vxi} \diff\vxi}\\
    		&\le\lim_{k\to+\infty}\norm{\fF^{-1}[\phi_{n_k}]-\fF^{-1}[\phi^*]}_{H^{\frac{\alpha}{2}}} \left(\int_{\sR^d}\frac{\Abs{\E^{2\pi\I\vx\vxi}}^2}{\langle\vxi\rangle^\alpha}\diff\vxi\right)^{\frac{1}{2}}\\
    		&=C\lim_{k\to+\infty}\norm{\fF^{-1}[\phi_{n_k}]-\fF^{-1}[\phi^*]}_{H^{\frac{\alpha}{2}}}=0.
    	\end{align*}
    	Hence $\fP_{\vX}\phi^*=\vY$ and $\phi^*\in \fA_{\vX,\vY}$.
    	
    	\noindent 7. 
    	Note that
    	\begin{equation*}
    		m=\inf_{\phi\in \overline{\fA_{\vX,\vY}}}\norm{\fF^{-1}[\phi]}_{H^{\frac{\alpha}{2}}}
    		\leq
    		\inf_{\phi\in \fA_{\vX,\vY}}\norm{\fF^{-1}[\phi]}_{H^{\frac{\alpha}{2}}}
    		\leq\norm{\fF^{-1}[\phi^*]}_{H^{\frac{\alpha}{2}}}
    		=m.
    	\end{equation*}
    	This implies that $\inf_{\phi\in \fA_{\vX,\vY}}\norm{\fF^{-1}[\phi]}_{H^{\frac{\alpha}{2}}}=m$ and
    	$
    	\phi^*\in\arg\min_{\phi\in \fA_{\vX,\vY}}\norm{\fF^{-1}[\phi]}_{H^{\frac{\alpha}{2}}}
    	$, which completes the proof.
    \end{proof}

\section{Details of numerical experiments}
    \subsection{Special Case: One Data Point in One Dimension}\label{case1}
        To simplify the problem, we start with a single point $X=0\in\sZ$ with the label $Y=2$. Denote $\phi_j = \phi(\xi_j)$ for $j\in\sZ$. We also assume that the function $\phi$ is an even function. Then according to the definition of $\fP_{\vX}$, we have the following problem:
        \begin{exam}[Problem~\ref{prob..VariationalPointCloud} with a particular discretization]  
        	\begin{align}
        		& \min_{\phi\in\sR^M} \sum_{j=1}^M(1+{j}^2\Delta\xi^2)^{\frac{\alpha}{2}}\Abs{\phi_j}^{2}, \\
        		& \mathrm{s.t.}\quad \sum_{j=1}^M\phi_j\Delta\xi = 1,
        	\end{align}
        \end{exam}
        \noindent where we further assume $\phi_0 = \phi(0) = 0$.  If we denote $\vphi = {(\phi_1, \phi_2, \ldots, \phi_M)}^{\T}$, $b = \frac{1}{\Delta\xi}$, $\mA = (1, 1, \ldots, 1)\in\sR^M$ and
        \begin{equation*}
        	\mGamma = \sqrt{\lambda}
        	\begin{pmatrix}
        		(1+1^2\Delta\xi^2)^{\frac{\alpha}{4}} & & & \\
        		& (1+2^2\Delta\xi^2)^{\frac{\alpha}{4}} & & \\
        		& & \ddots & \\
        		& & & (1+M^2\Delta\xi^2)^{\frac{\alpha}{4}}
        	\end{pmatrix}.
        \end{equation*}
        In fact this is a standard Tikhonov regularization~\citep{1977Solutions} also known as ridge regression problem with the Lagrange multiplier $\lambda$. The corresponding ridge regression problem is,
        \begin{equation}
        	\min_{\vphi}{\norm{\mA\vphi - b}_2^2 + \norm{\mGamma\vphi}_2^2},
        \end{equation}
        where we put $\lambda$ in the optimization term $\norm{\mGamma\vphi}_2^2$, instead of the constraint term $\norm{\mA\vphi - b}_2^2$. This problem admits an explicit and unique solution~\citep{1977Solutions},
        \begin{equation}\label{ridge_solution2}
        	\vphi = {(\mA^{\T}\mA + \mGamma^{\T}\mGamma)}^{-1}\mA^{\T} b.
        \end{equation}
        Here we need to point out that the above method is also applicable to the case that the matrix $\mGamma$ is not diagonal.
        
        Back to our problem, in order to obtain the explicit expression for the optimal $\vphi$ we need the following relation between the solution of the ridge regression and the singular-value decomposition (SVD).
        
        By denoting $\tilde{\mGamma} = \mI$ and
        \begin{equation*}
        	\tilde{\mA} = \mA\mGamma^{-1}
        	= \frac{1}{\sqrt{\lambda}}\left( (1+1^2\Delta\xi^2)^{\frac{\alpha}{4}}, (1+2^2\Delta\xi^2)^{\frac{\alpha}{4}}, \ldots, (1+M^2\Delta\xi^2)^{\frac{\alpha}{4}} \right),
        \end{equation*}
        where $\mI$ is the diagonal matrix, the optimal solution~\eqref{ridge_solution2} can be written as
        \begin{equation*}
        	\vphi = {(\mGamma^{\T})}^{-1}{\left( \tilde{\mA}^{\T}\tilde{\mA} +\mI \right)}^{-1}\mGamma^{-1}\mA^{\T} b
        	= {(\mGamma^{\T})}^{-1}{\left( \tilde{\mA}^{\T}\tilde{\mA} +\mI \right)}^{-1}\tilde{\mA}^{\T} b
        	= {(\mGamma^{\T})}^{-1}\tilde{\vphi},
        \end{equation*}
        where $\tilde{\vphi}={\left( \tilde{\mA}^{\T}\tilde{\mA} +\mI \right)}^{-1}\tilde{\mA}^{\T} b$ is the solution of ridge regression with $\tilde{\mA}$ and $\tilde{\mGamma}$. In order to obtain the explicit expression for $\tilde{\vphi}$ we need the following relation between the solution of the ridge regression and the singular-value decomposition (SVD).
        
        \begin{lem}\label{ridge_svd}  
        	If $\tilde\mGamma = \mI$, then this least-squares solution can be solved using SVD. Given the singular value decomposition
        	\begin{equation*}
        		\tilde\mA = \mU\mSigma\mV^{\T},
        	\end{equation*}
        	with singular values $\sigma_i$, the Tikhonov regularized solution can be expressed aspects
        	\begin{equation*}
        		\tilde\vphi = \mV\mD\mU^{\T} b,
        	\end{equation*}
        	where $\mD$ has diagonal values
        	\begin{equation*}
        		D_{ii} = \frac{\sigma_i}{\sigma_i^2 + 1},
        	\end{equation*}
        	and is zero elsewhere.
        \end{lem}
        \begin{proof} 
        	In fact, $\tilde\vphi={(\tilde\mA^{\T}\tilde\mA + \tilde\mGamma^{\T}\tilde\mGamma)}^{-1}\tilde\mA^{\T} b
        	=\mV(\mSigma^\T\mSigma+1\vI)^{-1}\mV^\T\mV\mSigma^\T\mU^\T b\\
        	=\mV\mD\mU^{\T} b$, which completes the proof.
        \end{proof}

        \noindent Since $\tilde{\mA}\tilde{\mA}^{\T} = \dfrac{1}{\lambda}\sum_{j = 1}^M (1+j^2\Delta\xi^2)^{-\frac{\alpha}{2}}$, we have $\tilde{\mA} = U \Sigma \vV^{\T}$ with
        \begin{equation*}
        	U = 1, \quad \Sigma = \frac{1}{\sqrt{\lambda}}{\left( \sum_{j = 1}^M (1+j^2\Delta\xi^2)^{-\frac{\alpha}{2}} \right)}^{\frac{1}{2}} := {Z}/{\sqrt{\lambda}},
        \end{equation*}
        \begin{equation*}
        	\vV = {\left( (1+1^2\Delta\xi^2)^{-\frac{\alpha}{2}}/Z, (1+2^2\Delta\xi^2)^{-\frac{\alpha}{2}}/Z, \ldots, (1+M^2\Delta\xi^2)^{-\frac{\alpha}{2}}/Z \right)}^{\T}.
        \end{equation*}
        Then we get the diagonal value
        \begin{equation*}
        	D = \frac{{Z}/{\sqrt{\lambda}}}{{Z^2}/{\lambda} + 1}.
        \end{equation*}
        Therefore, by Lemma ~\ref{ridge_svd}
        \begin{equation*}
        	\tilde{\vphi} = \vV D U b = \dfrac{{1}/{\sqrt{\lambda}}}{{Z^2}/{\lambda} + 1}{\left( (1+1^2\Delta\xi^2)^{-\frac{\alpha}{2}},(1+2^2\Delta\xi^2)^{-\frac{\alpha}{2}}, \ldots, (1+M^2\Delta\xi^2)^{-\frac{\alpha}{2}} \right)}^{\T} b.
        \end{equation*}
        Finally, for the original optimal solution
        \begin{align*}
        	\vphi   = {(\mGamma^{\T})}^{-1}\tilde{\vphi} =  \frac{1}{(Z^2 + \lambda)\Delta\xi}{\left( (1+1^2\Delta\xi^2)^{-\frac{\alpha}{2}}, (1+2^2\Delta\xi^2)^{-\frac{\alpha}{2}}, \ldots, (1+M^2\Delta\xi^2)^{-\frac{\alpha}{2}} \right)}^{\T},   
        \end{align*}
        which means
        \begin{equation*}
        	\phi_j = \frac{(1+j^2\Delta\xi^2)^{-\frac{\alpha}{2}}}{(Z^2 + \lambda)\Delta\xi}.
        \end{equation*}
        To derive the function in $x$ space, say $h(x)$, then
        \begin{align}\label{h(x)2}
        	h(x) & = \frac{1}{(Z^2 + \lambda)} \sum_{j=-M}^{M} (1+j^2\Delta\xi^2)^{-\frac{\alpha}{2}} \E^{2\pi\I j x} \nonumber\\
        	& = \frac{2}{(Z^2 + \lambda)} \sum_{j=1}^{M} (1+j^2\Delta\xi^2)^{-\frac{\alpha}{2}} \cos(2\pi j x).
        \end{align}
        
        Fig.~\ref{fig:hx_is_nontrivial}  shows that for this special case with a large $M$, $h(x)$ is not an trivial function in $\alpha>d$ case and degenerates to a trivial function in $\alpha<d$ case.

    
    \subsection{General Case: $n$ Points in $d$ Dimension}
        Assume that we have $n$ data points $\vx_1,\vx_2,\ldots,\vx_n\in \sR^d$ and each data point has $d$ components:
        \begin{equation*}
        	\vx_i=\left(x_{i1},x_{i2},\ldots,x_{id}\right)^\T
        \end{equation*}
        and denote the corresponding label as $\left(y_1,y_2,\ldots,y_n\right)^\T$. For the sake of simplicity, we denote the vector $(j_1,j_2,\cdots,j_d)^\T$ by $\vJ_{j_1\ldots j_d}$. Then our problem becomes
        \begin{exam}[Problem \ref{prob..VariationalPointCloud} with general discretization]
        	\begin{align}
        		& \min_{\phi\in \sR^{(2M)^d}} \sum_{j_1,\ldots,j_d=-M}^M(1+\norm{\vJ_{j_1\ldots j_d}}^2\Delta\xi^2)^{\frac{\alpha}{2}}\Abs{\phi_{j_1\ldots j_d}}^{2}, \\
        		& \mathrm{s.t.}\quad \sum_{j_1,\ldots,j_d=-M}^M\phi_{j_1\ldots j_d}\E^{2\pi\I\Delta\xi\vJ_{j_1\ldots j_d}^\T\vx_k}=y_k, \ \ k=1,2,\ldots,d
        	\end{align}
        \end{exam}
        
        \noindent The calculation of this example can be completed by the method analogous to the one used in subsection~\ref{case1}. Let 
        \begin{equation}\label{Aj}
        	\mA_j=\left(\E^{2\pi\I\Delta\xi\vJ_{-M-M\ldots -M}^\T\vx_j},\ldots, \E^{2\pi\I\Delta\xi\vJ_{j_1j_2\ldots j_d}^\T\vx_j},\ldots,\E^{2\pi\I\Delta\xi\vJ_{MM\ldots M}^\T\vx_j}\right)^\T,\ j=1,2,\ldots,n,
        \end{equation}

        \begin{equation}\label{A}
        	\mA=\left(\mA_1,\mA_2,\ldots,\mA_n \right)^\T\in \sR^{n\times (2M)^d},\quad
        	\vb=\left(y_1,y_2,\ldots,y_n\right)^\T\in \sR^{n\times 1},
        \end{equation}
        \begin{equation}\label{Gamma}
        	\mGamma=\lambda
        	\begin{pmatrix}
        		\ddots & &  \\
        		& (1+\norm{\vJ_{j_1j_2\ldots j_d}}^2\Delta\xi^2)^{\frac{\alpha}{4}}  & \\
        		& & \ddots 
        	\end{pmatrix}\in \sR^{(2M)^d\times(2M)^d}.
        \end{equation}
        We just need to solve the following equation:
        \begin{equation}\label{phi}
        	\vphi = {(\mA^{\T}\mA + \mGamma^{\T}\mGamma)}^{-1}\mA^{\T} b.
        \end{equation}
        Then we can obtain the output function $h(x)$ by using inverse Fourier transform:
        \begin{equation}\label{h}
        	h(\vx) = \sum_{j_1,\ldots,j_d=-M}^M\phi_{j_1\ldots j_d}\E^{2\pi\I\Delta\xi\vJ_{j_1\ldots j_d}\cdot\vx}
        \end{equation}
        Since the size of the matrix is too large, it is difficult to solve $\vphi$ by an explicit calculation. Thus we choose special $n$, $d$ and $M$ and show that $h(x)$ is not a trivial solution (non-zero function).
        
        In our experiment, we set the hyper-parameter $M,\alpha,\lambda,\Delta\xi$ in advance. We set $\lambda=0,5,\Delta\xi=0.1$ in 1-dimensional case and $\lambda=0.2,\Delta\xi=0.1$ in 2-dimensional case. We select two data points $\{(-0.5,0.9),(0.5,0.9)\}$ as the given points in 1-dimensional case and four points as given points in 2-dimensional case whose second coordinates are 0.5 so that it is convenient to observe the phenomenon. At first, we use formula \eqref{Aj}, \eqref{A} and \eqref{Gamma} to calculate matrix $\mA, \mGamma$ and vector $\vb$. Then from the equation \eqref{phi} we can deduce vector $\vphi$. The final output function $h(\vx)$ is obtained by inverse discrete Fourier transform \eqref{h}.
        
        In Fig.\ref{fig:diff_M}, we set $\alpha=10$ in both cases to ensure $\alpha>d$ and change the band limit $M$. We observe that as $M$ increases, the fitting curve converges to a non-trivial curve. In Fig.\ref{fig:diff_alpha}, we set $M=1000$ in 1-dimensional case and $M=100$ in 2-dimensional case. By changing exponent $\alpha$, we can see in all cases, the fitting curves are non-trivial when $\alpha>d$, but degenerate when $\alpha<d$. 

\end{document}